\documentclass{article}

\usepackage{arxiv}

\usepackage[utf8]{inputenc} % allow utf-8 input
\usepackage[T1]{fontenc}    % use 8-bit T1 fonts
\usepackage{hyperref}       % hyperlinks
\usepackage{url}            % simple URL typesetting
\usepackage{booktabs}       % professional-quality tables
\usepackage{amsfonts}       % blackboard math symbols
\usepackage{nicefrac}       % compact symbols for 1/2, etc.
\usepackage{microtype}      % microtypography
\usepackage{lipsum}		% Can be removed after putting your text content
\usepackage{graphicx}
\usepackage{natbib}
\usepackage{doi}
\usepackage{subcaption}
\usepackage{multirow} 
\usepackage{amsmath}
\usepackage{bm}

\title{Sleep-Disordered Breathing Detection with Quantum Recurrent Neural Networks}

\author{
    \href{https://orcid.org/0009-0007-7084-3678}{\includegraphics[scale=0.06]{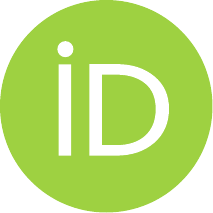}\hspace{1mm}Alejandro Mayorga-Redondo} \\
    Universidade da Coruña, CITIC\\
    Departamento de Ciencias de la Computación y \\ Tecnologías de la Información\\
    Campus de Elviña, 15071 A Coruña, Spain\\
    \texttt{alejandro.mayorga@udc.es}
    \And
    \href{https://orcid.org/0000-0002-0041-6802}{\includegraphics[scale=0.06]{orcid.pdf}\hspace{1mm}José Daniel Viqueira} \\
    CESGA, Galicia Supercomputing Center\\
    Universidade de Santiago de Compostela, COGRADE\\
    Departamento de Electrónica e Computación\\
    Campus Vida, 15782 Santiago de Compostela, Spain\\
    \texttt{jdviqueira@cesga.es}
    \And
    \href{https://orcid.org/0000-0001-5790-0577}{\includegraphics[scale=0.06]{orcid.pdf}\hspace{1mm}Diego Alvarez-Estevez} \\
    Universidade da Coruña, CITIC\\
    Departamento de Ciencias de la Computación y \\ Tecnologías de la Información\\
    Campus de Elviña, 15071 A Coruña, Spain\\
    \texttt{diego.alvareze@udc.es}
    \And
    \href{https://orcid.org/0000-0002-4894-1067}{\includegraphics[scale=0.06]{orcid.pdf}\hspace{1mm}Eduardo Mosqueira-Rey} \\
    Universidade da Coruña, CITIC\\
    Departamento de Ciencias de la Computación y \\ Tecnologías de la Información\\
    Campus de Elviña, 15071 A Coruña, Spain\\
    \texttt{eduardo.mosqueira@udc.es}
    \And
    \href{https://orcid.org/0000-0001-7272-8488}{\includegraphics[scale=0.06]{orcid.pdf}\hspace{1mm}Andrés Gómez} \\
    CESGA, Galicia Supercomputing Center\\
    15705 Santiago de Compostela, Spain\\
    \texttt{agomez@cesga.es}
}

\date{}

\renewcommand{\headeright}{}
\renewcommand{\undertitle}{}

\hypersetup{
pdftitle={Sleep-Disordered Breathing Detection with Quantum Recurrent Neural Networks},
pdfsubject={cs.LG, physics.quant-ph},
pdfauthor={Alejandro Mayorga-Redondo, José Daniel Viqueira, Diego Alvarez-Estevez, Eduardo Mosqueira-Rey, Andrés Gómez},
pdfkeywords={Machine Learning, Polysomnography, Quantum Machine Learning, Quantum Recurrent Neural Networks, Sleep Event Detection, Time Series Classification},
}

\begin{document}
\maketitle

\begin{abstract}
	This work investigates the application of Quantum Recurrent Neural Networks to the binary classification of sleep-disordered breathing events in time-series data, aiming to distinguish between apneic and normal breathing episodes. A variational classical--quantum architecture is proposed and evaluated against classical Recurrent Neural Network models. Experiments were conducted on multiple real-world datasets with varying levels of complexity to assess the strengths and limitations of the quantum approach. In the comparative analysis, both classical and quantum models showed comparable performance. The quantum model achieved superior performance on the smallest dataset, while yielding slightly weaker results in the most complex setting. However, the quantum approach exhibited notable strengths, such as competitive performance despite a drastically reduced parameter count and low-qubit regime, as well as good generalization capabilities. These results suggest that, despite current hardware and algorithmic limitations, variational quantum recurrent models represent a promising alternative for time-series classification as Quantum Computing technologies continue to mature.
\end{abstract}

% keywords can be removed

\section{Introduction}
\label{sec:introduction}

Sleep-related disorders represent a significant public health concern, affecting millions of individuals worldwide and increasing the risk of cardiovascular, metabolic, and neurological complications \citep{ramar2021sleep}. Their diagnosis commonly relies on polysomnography (PSG), the clinical gold standard for sleep assessment. PSG records multiple physiological variables throughout the night, including electroencephalography (EEG), electrocardiography (ECG), and respiratory signals, providing a comprehensive characterization of the patient's sleep activity \citep{bloch1997poly}. 

Among the events identified during polysomnography, breathing events such as apnea and hypopnea are of particular clinical relevance. According to the American Academy of Sleep Medicine, apnea is defined as a complete cessation of airflow lasting at least 10 seconds, whereas hypopnea corresponds to a partial reduction event \citep{berry2017AASM}. The identification of these events is crucial for the diagnosis of several sleep-breathing disorders. However, it requires trained specialists to manually inspect several hours of multichannel physiological recordings, making the diagnostic process labor-intensive, time-consuming, and subject to inter-scorer variability.

Machine Learning (ML) is a branch of artificial intelligence that focuses on learning patterns and relationships rather than relying on explicitly programmed rules \citep{jordan2015mloverview}. Its ability to process complex, high-dimensional data has motivated its use for automating sleep-recording analysis. Convolutional Neural Networks (CNNs) have demonstrated strong performance in extracting local temporal patterns from physiological signals, while Recurrent Neural Networks (RNNs) and their variants, including Long Short-Term Memory (LSTM) networks and Gated Recurrent Units (GRUs), are particularly effective at modeling long-range temporal dependencies \citep{siami2019performance,barnes2022,casal2021rnnoximetry}. Hybrid CNN--RNN architectures have further improved performance by combining spatial feature extraction with temporal modeling \citep{sun2019sleepstaging}. Despite these advances, classical ML models often require large amounts of labeled data, substantial computational resources, and a considerable number of trainable parameters to achieve high performance \citep{geiger2020dl}. These limitations motivate the exploration of alternative computational paradigms capable of learning complex temporal patterns more efficiently.

Quantum Computing (QC) has emerged as a fundamentally different computational paradigm that exploits the principles of quantum mechanics to process information. Its unique properties have motivated the exploration of quantum algorithms capable of solving certain computational problems more efficiently than their classical counterparts, particularly in optimization, simulation, cryptography, and large-scale linear algebra \citep{gill2021quantumcomputing}. Although current quantum hardware remains limited by noise, decoherence, and the relatively small number of available qubits, significant advances have been made, enabling the practical implementation of hybrid quantum-classical algorithms that combine the strengths of both computational paradigms \citep{endo2021hybrid}.

One of the most active research areas within this field is Quantum Machine Learning (QML), which investigates how quantum resources can be exploited to improve learning algorithms and data processing techniques \citep{schuld2015introduction}. Rather than replacing classical ML, many QML approaches integrate parameterized quantum circuits into conventional learning pipelines, where quantum and classical components are jointly optimized. This hybrid strategy is particularly well suited to current quantum hardware and has given rise to a variety of quantum models for classification, regression, and generative modeling tasks \citep{deLuca2021hybridQML}. Among these, Quantum Neural Networks (QNNs), especially those based on variational quantum circuits, have attracted considerable attention due to their expressive capabilities and compact parameterization. These characteristics make them particularly appealing for analyzing complex biomedical time-series data, where they have the potential to model intricate temporal relationships while requiring fewer trainable parameters than comparable classical architectures \citep{taha2018eeg}.

In this work, we investigate the use of Quantum Recurrent Neural Networks (QRNNs) for the binary classification of respiratory sleep events. Specifically, the proposed approach classifies segments of time-series data as either normal samples or disordered-breathing events. The objective is to evaluate the performance of a hybrid QRNN and compare it with an equivalent classical recurrent model. Beyond conventional classification metrics, the comparison focuses on several practical aspects, including parameter, computational and data efficiency under limited training data, as well as generalization capability on previously unseen samples.

The remainder of this work is organized as follows: Section~\ref{sec:background} introduces relevant concepts and the intersection of QC and ML. Section~\ref{sec:materials_and_methods} describes the datasets, the developed models, and the employed methodology. Section~\ref{sec:results_and_analysis} presents the experimental results and analyzes them. Finally, Section~\ref{sec:discussion_and_conclusions} discusses the findings and concludes the article, outlining future research directions.

\section{Background}
\label{sec:background}

In this section, we present the necessary theoretical background about Quantum Computing and particularly Quantum Machine Learning, needed to understand the remainder of the study.

\subsection{Quantum Computing \label{subsec:quantum_computing}}

QC represents a inherently distinct computing framework that harnesses the principles of quantum mechanics, utilizing quantum bits (qubits) rather than classical bits. Unlike classical bits, which can only take the values 0 or 1, qubits can exist in superpositions of both states and can be entangled with one another, creating states of the joint system with no classical analog. These inherently stochastic properties enable quantum algorithms to explore vast computational spaces, offering theoretical exponential or quadratic speedups for specific computational tasks over conventional classical computers, such as simulating complex physical systems, factoring large integers, or executing Monte Carlo methods \citep{wang_quantum_2012,nielsen_quantum_2011}.

To harness these quantum mechanical phenomena, researchers have developed several distinct physical and theoretical paradigms for quantum computation. One approach is quantum annealing, which is specifically tailored for combinatorial problems. This method finds the global minimum of a complex objective function by slowly evolving a known initial quantum state into the ground state of a target Hamiltonian \citep{arai_effectiveness_2023}. Other approaches include continuous-variable photonic QC, which encodes information in continuous degrees of freedom such as electromagnetic field amplitudes \citep{braunstein_quantum_2005}, and topological QC, which leverages topologically protected degenerate ground states to achieve intrinsic fault tolerance at the physical layer \citep{kitaev_fault-tolerant_2003}.

The gate-based (circuit) model is the paradigm we employ in this work, representing the prevailing approach toward achieving universal quantum computation. In this framework, computation proceeds by applying a sequence of discrete, reversible unitary operations, known as quantum logic gates, to a register of qubits, ending in measurements that yield classical outputs \citep{barenco1995elementary}. At present, gate-based implementations operate in the so-called Noise Intermediate-Scale Quantum (NISQ) regime, characterized by devices with tens to hundreds of qubits subject to physical decoherence and operational gate errors \citep{lau2022NISQ}. Despite these noise constraints, which prevent the execution of arbitrarily deep circuits, NISQ hardware has demonstrated its usefulness on programmable superconducting platforms, and useful algorithms have been developed that take these constraints into account \citep{arute_quantum_2019}.

\subsection{Quantum Machine Learning for sleep medicine}
\label{subsec:quantum_machine_learning}

Quantum Machine Learning applies the principles of QC to learning tasks, aiming to exploit quantum-mechanical phenomena to improve the representation and processing of complex data. A widely adopted approach is based on Variational Quantum Algorithms (VQAs), where parameterized quantum circuits are optimized through a hybrid quantum--classical training procedure  and later measured through quantum observables, such as expectation values of Pauli operators \citep{cerezo2021VQA,kohda2022quantum}. In these models, a classical optimizer iteratively updates the circuit parameters to minimize a cost function evaluated on a quantum processor, making them particularly suitable for current NISQ devices.

The application of QML to biomedical time-series analysis is especially appealing due to the high dimensionality, temporal dependencies, and complex correlations exhibited by physiological signals. Quantum models have the potential to encode high-dimensional feature spaces using relatively compact parameterizations, potentially improving parameter efficiency while reducing the number of trainable variables. Moreover, hybrid quantum--classical architectures can naturally integrate with existing ML pipelines, allowing quantum circuits to complement rather than replace classical feature extraction methods \citep{Maheshwari2022biomedical}.

Among proposed QML architectures, the Quantum Recurrent Neural Network is one of the most notable implementations. These models extend the memory-based processing of Classical Recurrent Neural Networks (CRNNs) into the quantum domain by retaining part of the quantum state across computational steps, enabling the modeling of temporal dependencies in sequential data \citep{li2023QRNN}. This characteristic makes QRNNs particularly well suited for physiological signals such as EEG, ECG, or respiratory recordings obtained during polysomnography.

Although QML remains an emerging field within sleep medicine, the first reported results are encouraging. QRNNs applied to EEG classification have achieved high accuracy, demonstrating performance comparable to classical recurrent models \citep{taha2018eeg}. Similarly, hybrid quantum-classical models for automatic sleep stage classification have reported competitive results while requiring fewer trainable parameters than purely classical approaches \citep{tian2025qmlstaging}. Beyond sleep medicine, QML has also been successfully explored in related biomedical applications, including heart disease diagnosis \citep{pitchal2024QCNNheart} and dementia classification \citep{amin2025QCNNdementia}. These studies suggest that quantum learning models constitute a promising alternative for the analysis of complex medical time series, although their practical advantages over classical approaches remain an active area of research.

\section{Materials and methods}
\label{sec:materials_and_methods}

In this section, we describe the datasets, models, and methodology used to obtain the presented results.

\subsection{Dataset description}
\label{sec:dataset}

To evaluate model performance under different conditions, three datasets with varying characteristics were defined. They were derived from seven PSG recordings from the Sleep Heart Health Study follow-up phase database (SHHS-2). This study, implemented by the National Heart, Lung, and Blood Institute, was designed with the objective of analyzing the cardiovascular and other health consequences of sleep-disordered breathing among middle-aged patients. The complete database is available upon request at the National Sleep Research Resource \citep{zhang2018national}.

Each of the three derived datasets consists of a collection of 60-second windows of physiological signals related to the patient's breathing, namely the \textit{airflow} signal, which measures respiratory flow, and the \textit{SaO\textsubscript{2}} signal, which measures blood oxygen saturation. The task is formulated as a balanced binary classification problem between apneic (apnea or hypopnea) events and normal samples. To standardize the temporal resolution, all signals were resampled to 2 Hz. \textit{SaO\textsubscript{2}} was resampled using linear interpolation, while \textit{airflow} was resampled using a Fast Fourier Transform-based method. This resulted in instances of 120 timesteps per signal. In addition, periods containing artifacts or corresponding to wakefulness states were removed to ensure signal quality and consistency across all datasets. Since \textit{SaO\textsubscript{2}} represents the percentage of oxygen saturation in blood and therefore ranges from 0 to 100\%, values below 75\% were clipped under the premise that they fall outside the physiologically expected range and do not provide relevant information for the task. In contrast, the \textit{airflow} signal has no specific physical units and was provided already normalized to the range [-1, 1]. Subsequently, \textit{SaO\textsubscript{2}} was normalized per-recording to [-1, 1] to match the scale of the \textit{airflow} signal. Per-recording rather than per-instance normalization was used to preserve the relative magnitude of oxygen desaturation events across patients, an important indicator of respiratory events. Finally, for apneic instances, a random temporal offset was introduced so that apnea or hypopnea events could begin at any position within the instance while ensuring that the complete annotated event was contained.

Despite these common preprocessing steps, the datasets differ in both size and composition, allowing the proposed models to be evaluated under different levels of complexity. Their names (small, medium and large dataset) refer to the total number of available instances. The small dataset contains only the \textit{airflow} signal (number of input variables $n_v=1$), whereas the medium and large datasets additionally include the \textit{SaO\textsubscript{2}} signal ($n_v=2$). Furthermore, the small and medium datasets contain only apnea events, while the large dataset includes both apnea and hypopnea events, producing a more challenging classification problem. A summary of the characteristics of each dataset is presented in Table \ref{tab:datasets_chars}.

\begin{figure}[h]
    \centering

    \begin{subfigure}{0.9\textwidth}
        \centering
        \includegraphics[width=\linewidth]{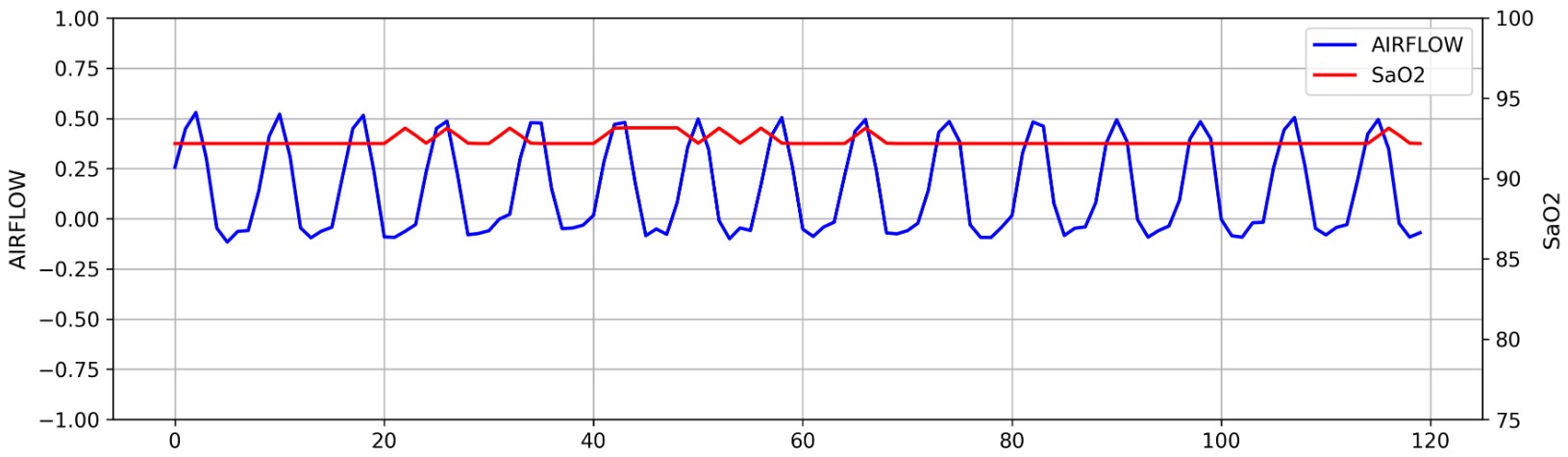}
        \caption{Normal instance}
        \label{fig:normal}
    \end{subfigure}

    \vspace{0.25cm}

    \begin{subfigure}{0.9\textwidth}
        \centering
        \includegraphics[width=\linewidth]{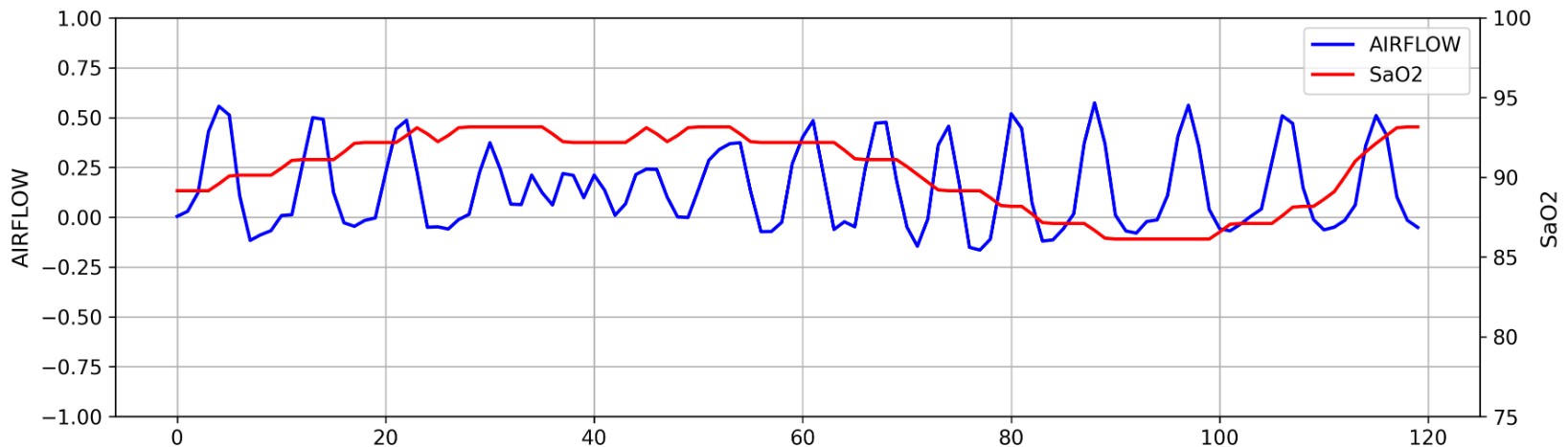}
        \caption{Hypopnea instance}
        \label{fig:hypopnea}
    \end{subfigure}

    \vspace{0.25cm}
        
    \begin{subfigure}{0.9\textwidth}
        \centering
        \includegraphics[width=\linewidth]{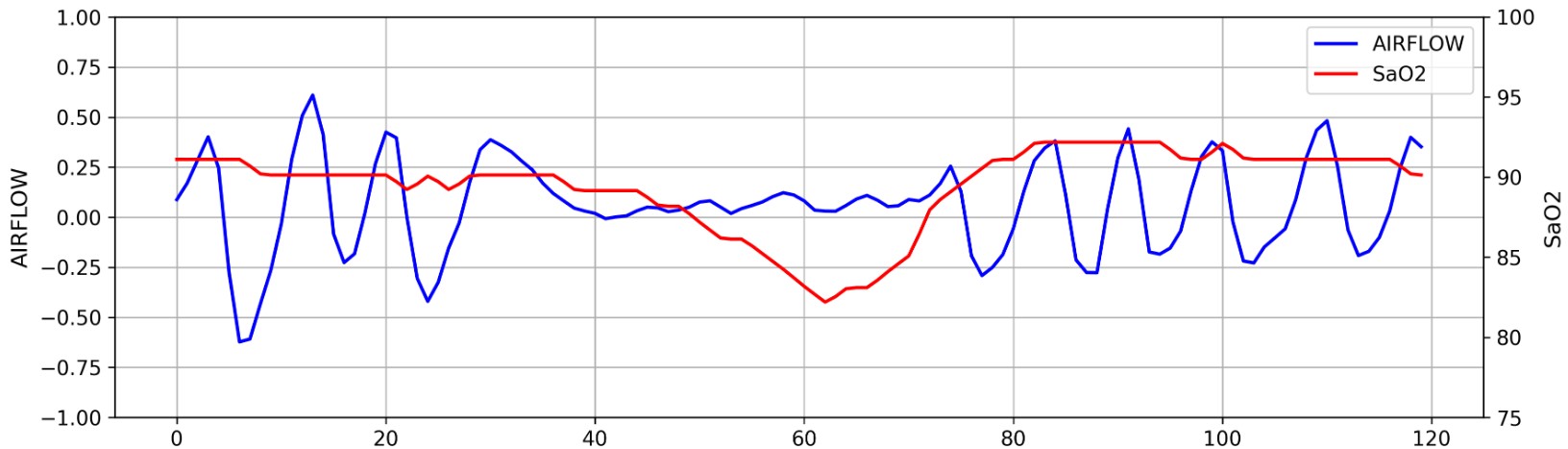}
        \caption{Apnea instance}
        \label{fig:apnea}
    \end{subfigure}

    \caption{Examples of \textit{airflow} and \textit{SaO\textsubscript{2}} signals for the different samples that compose the datasets. The X-axis represents timesteps, while the Y-axis indicates the corresponding amplitude for each signal.}
    \label{fig:instances}
\end{figure}

Figures \ref{fig:normal}, \ref{fig:hypopnea} and \ref{fig:apnea} illustrate representative normal, hypopnea and apnea samples according to the aforementioned specifications. Notable differences can be observed in both signals across the three instances. In the normal samples, the \textit{airflow} signal exhibits regular oscillations corresponding to stable breathing, while the \textit{SaO\textsubscript{2}} signal remains stable. In the hypopnea, the \textit{airflow} signal shows a partial reduction in respiratory amplitude approximately from timestep 30 to 60, accompanied by a decrease in \textit{SaO\textsubscript{2}} toward the end of the window. Finally, in the apnea, \textit{airflow} is almost completely interrupted from approximately timestep 40 to 70, resulting in an abrupt decrease in the \textit{SaO\textsubscript{2}} signal around the middle of the window.

\begin{table}
    \centering
    \begin{tabular}{@{}lcccc@{}}
    \toprule
     \textbf{Data features} 
     & Small dataset & Medium dataset & Large dataset &\\
    \midrule
          Instances number & 300 & 600 & 1200 \\
         Apneic events & Only apnea & Only apnea & Apnea and hypopnea \\
         Data distribution & 50/50/$-$ & 50/50/$-$ & 50/25/25 \\
         Used signals & \textit{airflow} & \textit{airflow} and \textit{SaO\textsubscript{2}} & \textit{airflow} and \textit{SaO\textsubscript{2}}\\
                 
    \hline
    \end{tabular}
    \caption{Employed datasets with their differential characteristics. Data distribution row reports, in order, the percentage of normal, apnea, and (when included) hypopnea instances.}
    \label{tab:datasets_chars}
\end{table}

\subsection{Recurrent Neural Network classifiers \label{subsec:RNN_class}}
\begin{figure}
    \centering
    \includegraphics[width=\linewidth]{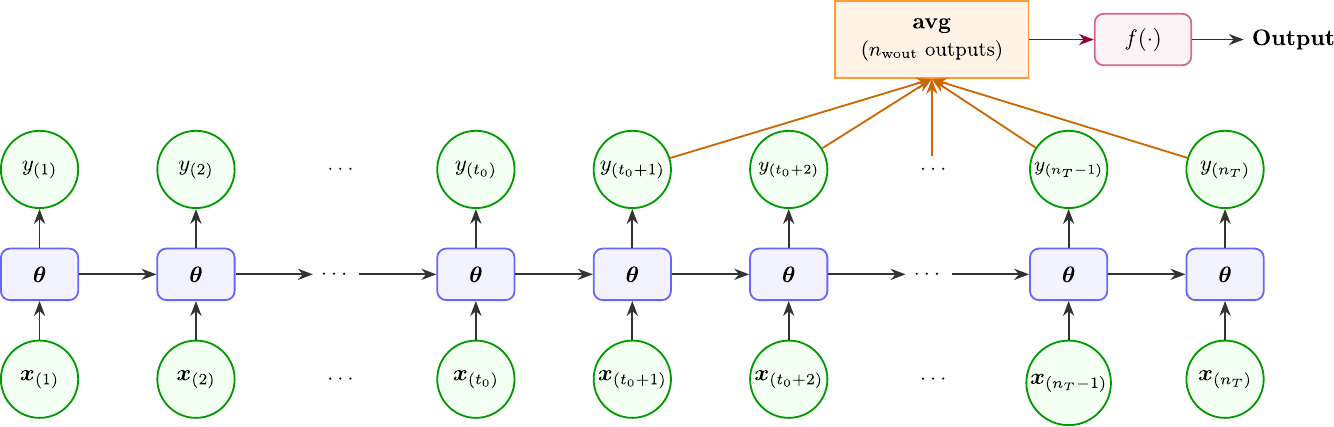}
    \caption{Diagram of the RNN classifier. We show a recurrent neural network unrolled through time. Each timestep, it processes an input vector, returns an output scalar and transmits memory to the next timestep. The first $t_0 = n_T - n_\text{wout}$ outputs are discarded and the last $n_\text{wout}$ enter the averaging phase. After averaging, a function $f$ returns an output value to classify the sample.}
    \label{fig:rnn_classifier}
\end{figure}

We define a classifier that, given a time-series instance of length $n_T$, $\left. \bm{x}_{(t)} \right|_{t=1}^{n_T}$, returns an output $y \in [0,1]$, interpreted as the probability of normal breathing (class 0) versus an apneic event (class 1). Figure~\ref{fig:rnn_classifier} illustrates the architecture, showing a recurrent structure unrolled through time, an average pooling function on a final subset of output nodes (timestep scalar outputs) and a function giving the final output for an instance.

The recurrent structure, which is the first and core component, varies across models. The classical models incorporate a simple RNN structure, with single hidden states. Meanwhile, the quantum models are evaluated through a QRNN circuit with intermediate measurements. The second component takes the last $n_\text{wout}$ nodes and averages them in both cases. An initial warm-up period is introduced by discarding the first output nodes before the classification stage. Although these initial timesteps are excluded from the final decision, they still contribute to the generation of the subsequent recurrent hidden states, allowing the network to accumulate temporal context before generating predictions. The third component is a real-valued function $f: \mathbb{R} \to \mathbb{R}$, which can be either a standard activation function or any general scalar-to-scalar mapping.

\subsubsection{Classical Recurrent Neural Network \label{subsubsec:CRNN}}

A classical recurrent architecture based on a vanilla CRNN was used as the baseline model in this study. The selection of this architecture is motivated by its structural similarity to the QRNN used here, allowing for a more consistent comparison between classical and quantum approaches. Since both models rely on recurrent processing of temporal information, the differences observed in performance can be more directly attributed to the underlying computational paradigm rather than to substantial architectural differences.

The proposed architecture consists of a stack of recurrent layers, each receiving the complete temporal sequence generated by the previous layer and progressively refining its representation. The number of recurrent units is gradually reduced from one layer to the next according to a base reduction hyperparameter, allowing the network to progressively compress the temporal representation while retaining the most informative features. Unlike conventional recurrent classifiers, which typically obtain outputs using only the last hidden state, the proposed model employs most of the hidden representations associated with every timestep for the final decision.

The remaining hidden representations are subsequently rearranged so that each temporal output is independently processed by a shared fully connected layer with a single neuron and sigmoid activation, producing a probability estimate for every considered timestep. Rather than selecting a single prediction, the architecture aggregates $n_\text{wout}$ timestep outputs through a Global Average Pooling layer plus a sigmoid activation function, yielding a single probability that constitutes the final binary classification output. To ensure a fair comparison with the quantum models, no regularization techniques, such as dropout or weight regularization, were incorporated into the architecture, thereby preventing performance differences arising from model-specific optimization strategies.

\subsubsection{Quantum Recurrent Neural Network}
\label{subsubsec:QRNN}

\begin{figure}
    \centering
    \begin{subfigure}{0.9\textwidth}
    \centering
    \includegraphics[scale=0.8]{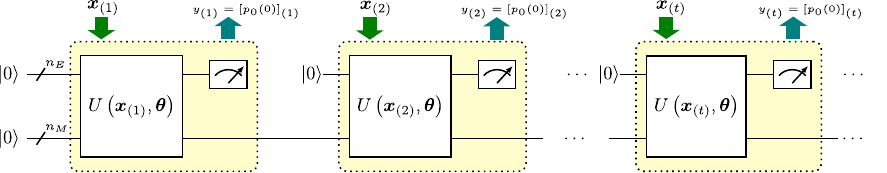}
    \caption{QRNN circuit representation. Input and output connections are plotted for a better understanding. Outputs are obtained through the probability of measuring 0 in the first qubit, $p_0(0)$.}
    \end{subfigure}
    
    \begin{subfigure}{0.90\textwidth}
    \centering
    \includegraphics[scale=0.8]{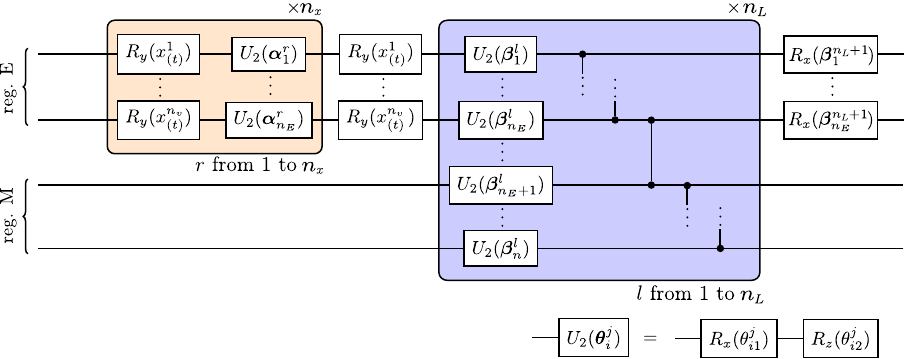}
    \caption{Inner structure of the operator $U(\bm{x}_{(t)}, \bm{\theta})$ used in this article.}
    \end{subfigure}
    
    \caption{Detailed QRNN architecture.}
    \label{fig:qrnn_classifier}
\end{figure}

The QRNN architecture, previously proposed and validated for multivariate time series prediction \citep{viqueira2025density}, processes sequential data through a parameterized quantum circuit (Figure~\ref{fig:qrnn_classifier}). The circuit is partitioned into two distinct components: an exchange register (E) comprising $n_E$ qubits, and a memory register (M) comprising the remaining $n_M$ qubits. Similar to classical recurrent models, the quantum state is evolved at each timestep using a unitary operator $U(\bm{x}_{(t)}, \bm{\theta})$, with the same set of trainable parameters $\bm{\theta}$ shared across the entire sequence.

During a single timestep, the input vector $\bm{x}_{(t)}$ is first loaded into the exchange register E via a data re-uploading operator, with one vector component parameterized into each of the $n_E$ qubits. To enhance the model's expressivity, this encoding block is repeated $n_x$ times \citep{perez-salinas2020data}. Following encoding, the full quantum state (registers E and M) is evolved using a hardware-efficient ansatz \citep{kandala_hardware-efficient_2017}. This evolving operator consists of $n_L$ layers of rotation gates and ladders of controlled-Z gates, serving a dual purpose: it entangles both registers while driving the state evolution through the circuit.

The number of layers $n_L$ in the evolving operator is a critical design choice constrained by the causal light cone of the circuit. To ensure that the state and operations on the last qubit can causally influence the expectation value of the first qubit \citep{braccia_computing_2024,leone_practical_2024}, the light cone must span all qubits. We therefore enforce a minimum depth of $n_L \geq n_E + n_M$. While adding further layers increases the capacity to explore the Hilbert space and improves expressivity, it also risks introducing vanishing-gradient problems and higher computational costs \citep{cho2026vanishing}. Consequently, we define the excess number of layers as a hyperparameter, $\Delta n_L \equiv n_L - (n_E + n_M)$.

At the end of each timestep, the exchange register E is measured and subsequently reset, while the memory register M is preserved to carry context into the next step. To process the measurement outcomes for binary classification, we diverge from prior work that relied on the global expectation value $\langle Z ^{\otimes n_E} \rangle$. Instead, we use the probability of measuring the $\vert{}0\rangle$ state in the first qubit, $p_0(0)$. This single-qubit probability naturally maps to the likelihood of classifying a sample as either 0 or 1. Thus, for a series of length $n_T$, this process yields $n_T$ intermediate probability measurements used for classification.

After the QRNN processes $n_T$ timesteps, the classifier averages $n_\text{wout}$ outputs to produce the final output for classification. $f$ is the identity function in this case, because the quantum model does not include activation functions.

In this study, the QRNN is emulated through a purpose-built emulator described in \citep{viqueira2025density}. Hardware and shot noise were not simulated; thus, the measurement probabilities were obtained through emulation with high precision and automatic differentiation was used with JAX \citep{bradbury2018jax}.

\subsection{Experimental setup}
\label{sec:experimental_setup}

To identify the optimal hyperparameter configuration, a grid search strategy was conducted for both the classical and quantum paradigms. The search focused on the most sensitive hyperparameters of each model, including those directly related to the recurrent architecture and the training procedure. This exploration allowed the evaluation of multiple parameter combinations in order to determine the configuration yielding the best overall performance for each dataset. A stratified 20\% split was held out as the test set after the grid search.

The hyperparameter optimization process was carried out using stratified 5-fold cross-validation. For every candidate configuration, the available training data were partitioned into five subsets, where four folds were used for training and the remaining fold was employed for validation. To obtain statistically robust results and account for variability induced by random parameter (weight) initialization, we set the number of repetitions to 5. Thus, 25 training runs were performed per hyperparameter configuration (5 folds $\times$ 5 repetitions). The performance metric was averaged across all repetitions. This iterative procedure ensured that each fold was used as validation data multiple times, leading to a more reliable estimate of model performance and reducing the risk of biased or overly optimistic results.

The networks were trained using the binary cross-entropy loss function, while the Adam optimizer was employed to update the trainable parameters owing to its adaptive learning-rate mechanism and its robust convergence properties \citep{tong2022ADAM}. Training was performed for a maximum of 500 epochs, incorporating an early stopping strategy based solely on training accuracy. This prevented the optimization process from systematically reaching the maximum number of epochs, greatly reducing overall training time.

When selecting the hyperparameter values, both architectural and learning-related parameters were taken into account. An effort was made to maintain a similar number of configurations in the grid search procedures for both the classical and quantum models in order to ensure a fair comparison between them. The specific values considered for each hyperparameter are detailed in Table \ref{tab:hparameters}. The search ranges were determined according to usual limits used for each kind of model. While classical RNN families typically require deeper architectures \citep{karim_2019_insights}, state-of-the-art QNNs often use only on the order of 1 to 10 layers \citep{lins2024qmldrowsiness}. This is mainly due to very short limits in emulation and the lack of quantum hardware to deploy the models.

After completing the grid search procedure, the hyperparameter configurations achieving the highest mean validation accuracy on each dataset were selected for the testing phase. The corresponding models were retrained using the complete training sets, reserving 16\% of the data as a validation set, matching the validation proportion employed during the cross-validation stage. During this final training, early stopping was applied by monitoring the validation loss with a patience of 10 epochs and a tolerance of 0.001 to reduce the risk of overfitting. Finally, the trained models were evaluated on the previously held-out test sets to obtain the performance metrics. This procedure ensures an unbiased assessment of the final models while preventing information leakage between the hyperparameter optimization and evaluation stages.

\begin{table}
   \centering
\begin{tabular}{@{}l@{\hspace{20pt}}lccc@{}}
    \toprule
         \textbf{Parameter type} &
         \textbf{Hyperparameter} &
         \textbf{Fixed values} &
         \textbf{Search values} \\
    \hline
         \multirow{7}{*}{Shared}
         & epochs & 500                    &                   &  \\
         & early stopping (acc) & 10 &  &  \\
         & optimizer & Adam & & \\
         & loss function & Binary cross-entropy & & \\
         & $n_\text{wout}$ & $n_T-20$ & & \\ & batch size  &               &  $\{16, 32,64\}$ \\
         & learning rate  &            & $\{0.0001,0.001\}$\\
         \midrule
         \multirow{4}{*}{CRNN}      & bias & True & & \\ & bidirectional & False & & \\
         & base reduction & & $\{2, 3, 4\}$ &  \\
         & CRNN layers & & $\{1, 2, 3, 4\}$ & \\
         & CRNN units & & $\{64, 128, 256\}$ & \\
         \midrule
         \multirow{5}{*}{QRNN}
         & qubits E ($n_E$) & $n_v$ & & \\
         & qubits M ($n_M$) & &$\{1,2,3,4,5\}$ \\
         & additional layers ($\Delta n_L$) & & $\{0,1\}$ \\
         & re-uploading number ($n_x$) & & $\{2,4,6\}$ \\
    \hline
    \end{tabular}
    \caption{Employed models and training hyperparameters. Each horizontal line divides the shared values with those applying only to one model. Sets in brackets define those implied in the grid search, while single values are fixed before.}
    \label{tab:hparameters}
\end{table}

\section{Results and analysis}
\label{sec:results_and_analysis}

The metrics obtained by the models, as well as the analysis of their results, are explained in this section.

\subsection{Summary of obtained metrics}

For each dataset in the grid search stage, the model with the highest mean accuracy across all folds and repetitions was selected as the best-performing configuration. Table \ref{tab:hparameters_performance} reports the average accuracy for each dataset, along with the standard deviation and the corresponding hyperparameter settings for both the classical and quantum models. Using these best configurations, the test set was then employed to compute the final evaluation metrics during the retraining phase, and the results are presented in Table \ref{tab:metrics_test}.

\begin{table}[h]
    \centering
    \label{tab:classification_results}
    \begin{tabular}{lcccccc}
        \toprule
        & \multicolumn{2}{c}{Small dataset} & \multicolumn{2}{c}{Medium dataset} & \multicolumn{2}{c}{Large dataset} \\
        \cmidrule(lr){2-3} \cmidrule(lr){4-5} \cmidrule(lr){6-7}
        Model & CRNN & QRNN & CRNN & QRNN & CRNN & QRNN \\
        \midrule
        folds mean acc.  & 0.6541 & 0.7225 & 0.8687 & 0.8717 & 0.8200 & 0.7702 \\
        folds mean std. dev.  & 0.0664 & 0.1152 & 0.0358 & 0.0306 & 0.0292 & 0.0469 \\
        \addlinespace
        batch size & 16 & 16 & 16 & 32 & 16 & 16 \\
        learning rate & 0.001 & 0.001 & 0.001 & 0.001 & 0.001 & 0.001 \\ 
        \addlinespace
        CRNN layers & 4 & $\_$ & 4 & $\_$ & 4 & $\_$ \\
        CRNN units  & 256 & $\_$ & 256 & $\_$ & 64 & $\_$ \\
        base reduction & 4 & $\_$ & 4 & $\_$ & 2 & $\_$ \\
        \addlinespace
        qubits M ($n_M$)  & $\_$ & 5 & $\_$ & 5 & $\_$ & 2 \\
        additional layers ($\Delta n_L$)  & $\_$ & 1 & $\_$ & 1 & $\_$ & 0 \\
        re-uploading number ($n_x$)  & $\_$ & 4 & $\_$ & 6 & $\_$ & 4 \\
        \bottomrule
    \end{tabular}
    \caption{Best models, their hyperparameter configurations, and their performance in 5-repeated 5-fold cross-validation.}
    \label{tab:hparameters_performance}
\end{table}

\begin{table}[h]
    \centering
    \begin{tabular}{llcccccc}
        \toprule
        & & \multicolumn{2}{c}{Small dataset} &
            \multicolumn{2}{c}{Medium dataset} &
            \multicolumn{2}{c}{Large dataset} \\
        \cmidrule(lr){3-4} \cmidrule(lr){5-6} \cmidrule(lr){7-8}
        Metric & Value & CRNN & QRNN & CRNN & QRNN & CRNN & QRNN \\
        \midrule

        \multirow{2}{*}{Accuracy}
        & Average  & 0.6900 & 0.8300 & 0.8850 & 0.8783 & 0.8033 & 0.7367 \\
        & std. dev.& 0.0898 & 0.0609 & 0.0370 & 0.0113 & 0.0448 & 0.0301 \\
        \addlinespace

        \multirow{2}{*}{F1-score}
        & Average  & 0.7130 & 0.8322 & 0.8775 & 0.8734 & 0.7935 & 0.7139 \\
        & Std. dev.& 0.0539 & 0.0605 & 0.0405 & 0.0157 & 0.0418 & 0.0263 \\
        \addlinespace

        \multirow{2}{*}{Precision}
        & Average  & 0.6957 & 0.8417 & 0.9362 & 0.9089 & 0.8420 & 0.7855 \\
        & Std. dev.& 0.1148 & 0.1157 & 0.0406 & 0.0270 & 0.0658 & 0.0512 \\
        \addlinespace

        \multirow{2}{*}{Recall}
        & Average  & 0.7533 & 0.8467 & 0.8266 & 0.8433 & 0.7516 & 0.6550 \\
        & Std. dev.& 0.0805 & 0.1087 & 0.0466 & 0.0478 & 0.0285 & 0.0100 \\
        \addlinespace

        \multirow{2}{*}{AUC-ROC}
        & Average  & 0.7586 & 0.8421 & 0.9439 & 0.9576 & 0.8555 & 0.8191 \\
        & Std. dev.& 0.1080 & 0.0601 & 0.0190 & 0.0037 & 0.0514 & 0.0350 \\
        \addlinespace

        \multirow{2}{*}{Epochs}
        & Average  & 22.60 & 245.60 & 46.60 & 254.80 & 39.60 & 159.60 \\
        & Std. dev.&  8.08 & 111.82 & 15.75 & 118.16 & 10.44 & 90.13 \\

        \bottomrule
    \end{tabular}
    \caption{Classification performance on the test set. Results are reported as averages over 5 runs and their corresponding standard deviations.}
    \label{tab:metrics_test}
\end{table}

For the test results, we use standard classification metrics to evaluate model performance in a binary classification setting. These metrics include accuracy, precision, recall, and F1-score, computed using a fixed classification threshold of 0.5. In addition, the Area Under the Receiver Operating Characteristic Curve (AUC-ROC) is reported as a threshold-independent measure of the model’s overall discriminative ability. The mean and standard deviation of all metrics across the five independent training runs are presented. The number of epochs is also reported, as it provides information about the time required for the models to converge, as well as the variability in training duration across executions.

\subsection{Performance evaluation}

The comparison of performance between classical and quantum approaches, shown in Table \ref{tab:metrics_test}, reveals that their relative performance strongly depends on both data availability and task complexity. The quantum model shows a clear advantage in the small dataset, outperforming the classical counterpart across all metrics and improving accuracy from 69\% to 83\%, which clearly indicates better generalization capabilities under limited training conditions. As more data become available, the performance gap narrows, with both models achieving similar accuracy and F1-score values in the medium dataset, although with different strengths: the classical model provides higher precision, whereas the quantum model achieves better recall and AUC-ROC values. However, this advantage does not extend to the large dataset, where the classical model consistently outperforms the quantum implementation across all metrics. Overall, these findings suggest that quantum models can provide competitive performance and even outperform classical methods in low-data scenarios, while classical approaches remain more robust as the complexity of the problem increases.

Although the small dataset contains a limited number of training instances, the models achieve noticeably better performance in the final evaluation than during the grid search. This improvement is mainly attributable to the retraining stage, where different random seeds and parameter initializations are used. Given the relatively small number of available samples, both the initial weights and the specific split between training and validation subsets have a considerable impact on the learned model, which is also reflected in the comparatively large standard deviations. Conversely, the large dataset represents a challenging classification problem due to the inclusion of hypopnea events. Since hypopneas constitute intermediate respiratory events between normal breathing and complete apneas, they blur the decision boundary between classes and thus reduce classification performance for both paradigms.

Regarding the selected hyperparameters, an interesting trend can be observed across the different datasets. For both the small and medium datasets, the best-performing configurations correspond to some of the largest model-specific architectures explored during the grid search. In particular, the classical RNN achieved its highest cross-validation accuracy using the maximum number of recurrent layers, recurrent units, and base reduction factor, while the QRNN selected the highest values for the number of memory qubits ($M$), additional layers ($\Delta n_L$), and data re-uploading number ($n_x$). This suggests that, for relatively well-defined classification problems, increasing the representational capacity of the models enables them to learn more discriminative features and capture more complex relationships within the data. In contrast, the large dataset favored considerably simpler configurations in both paradigms. A likely explanation, as noted above, is that this dataset poses a substantially more challenging classification problem due to the inclusion of hypopnea events, which blur the distinction between the two classes. Under these conditions, larger models may be more prone to fitting subtle patterns or noise that do not generalize well, whereas more compact architectures provide a better balance between model capacity and generalization. This observation is consistent across both the classical and quantum approaches, suggesting that the complexity of the classification task, rather than the learning paradigm itself, largely determines the optimal model size.

The average number of epochs required for convergence provides further insight into the retraining dynamics. For classical models, this metric remains relatively stable across all three datasets, typically ranging from 23 to 47 epochs. Notably, the medium dataset exhibits a significantly higher standard deviation than both the small and large datasets, suggesting greater variability in training behavior at this scale. In contrast, quantum models display a much broader convergence window, ranging from 71 to 464 epochs, with a standard deviation approximately ten times larger than that of classical models. This points to markedly slower and less consistent convergence in the quantum case.

\begin{figure}[h]
    \centering

    \begin{subfigure}{0.49\textwidth}
        \centering
        \includegraphics[width=\textwidth]{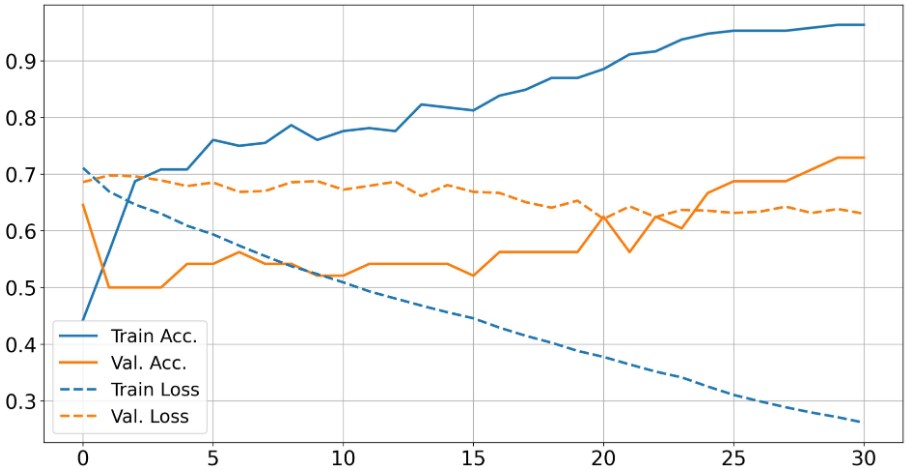}
        \caption{Classical model - small dataset}
        \label{fig:convergence_curves_small_dataset_classical_model}
    \end{subfigure}
    \hfill
    \begin{subfigure}{0.49\textwidth}
        \centering
        \includegraphics[width=\textwidth]{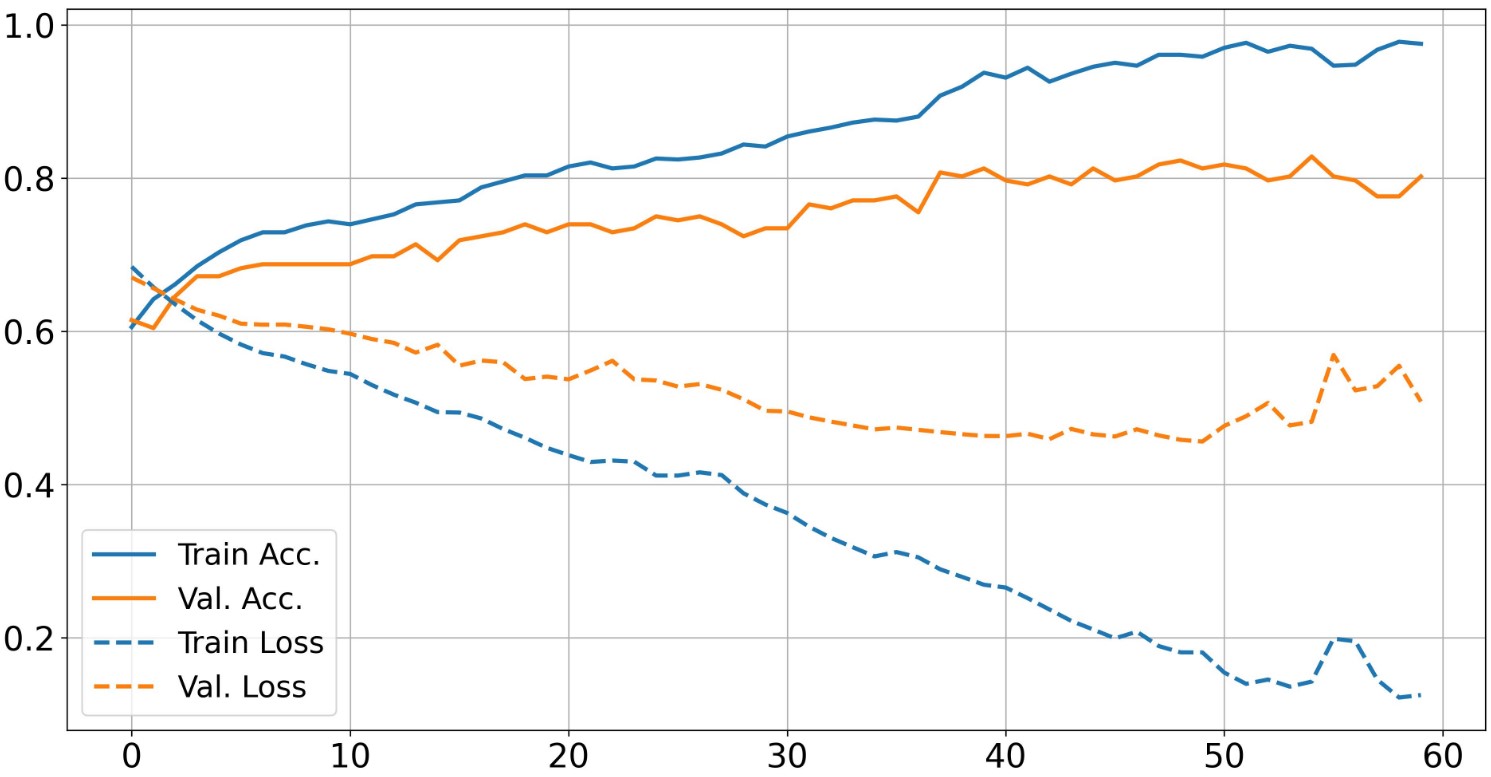}
        \caption{Classical model - large dataset}
        \label{fig:convergence_curves_large_dataset_classical_model}
    \end{subfigure}

    \vspace{0.5cm}

    \begin{subfigure}{0.49\textwidth}
        \centering
        \includegraphics[width=\textwidth]{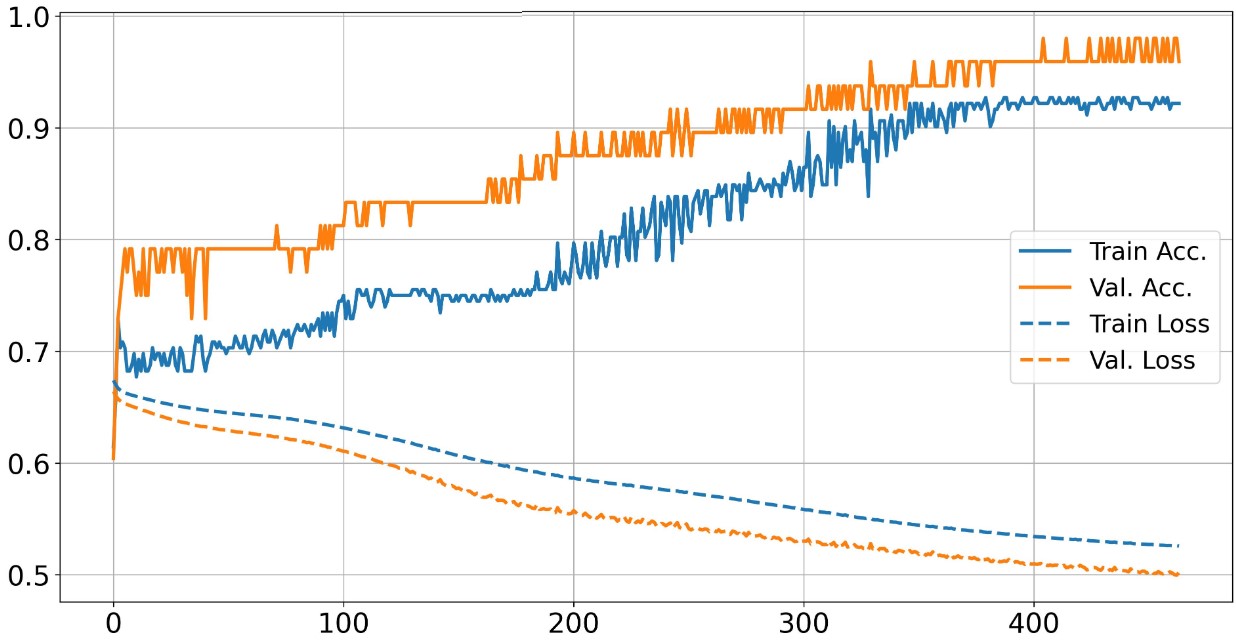}
        \caption{Quantum model - small dataset}
        \label{fig:convergence_curves_small_dataset_quantum_model}
    \end{subfigure}
    \hfill
    \begin{subfigure}{0.49\textwidth}
        \centering
        \includegraphics[width=\textwidth]{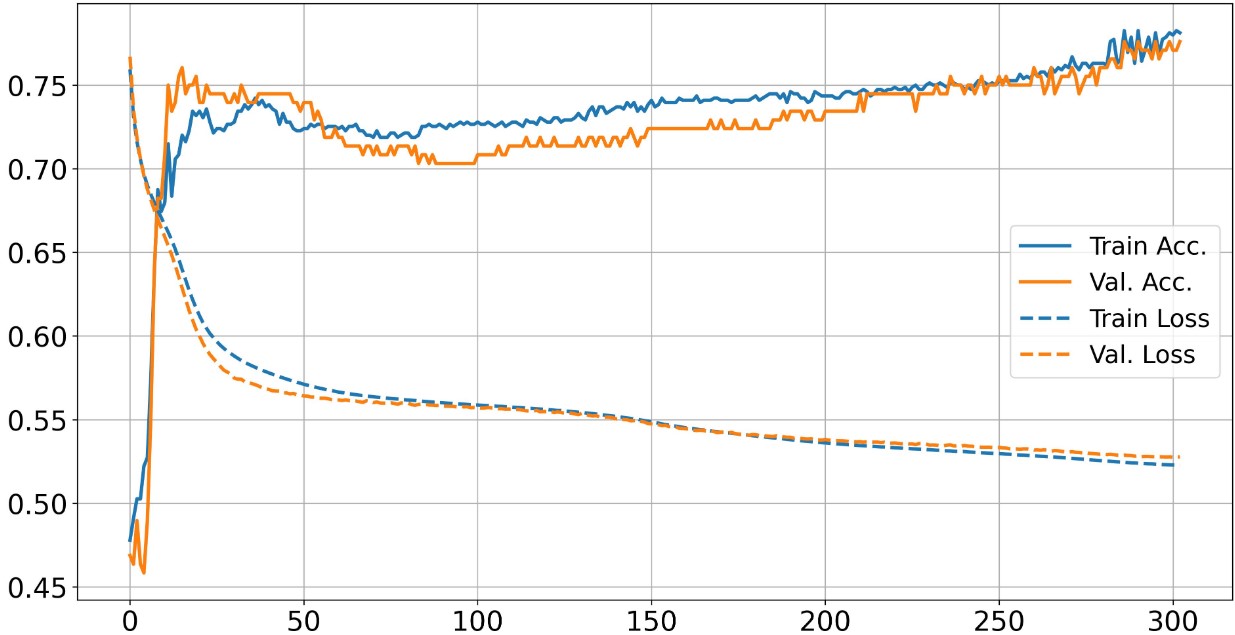}
        \caption{Quantum model - large dataset}
        \label{fig:convergence_curves_large_dataset_quantum_model}
    \end{subfigure}

    \caption{Training loss and accuracy convergence curves for the classical and quantum models. X-axis represents the number of epochs and Y-axis indicates the metric values.}
    \label{fig:convergence_curves_classical_quantum_models}
\end{figure}

Finally, Figure \ref{fig:convergence_curves_classical_quantum_models} presents the convergence curves obtained from one of the retraining runs for the small and large datasets, for both classical and quantum models. The selected retraining cases correspond to those requiring the highest number of epochs, providing a representative scenario for analyzing slower convergence behavior. For the classical model on the large dataset, the training and validation loss and accuracy curves follow relatively typical trends, suggesting adequate, though not perfect, generalization. In contrast, the validation curves for the small dataset show limited improvement compared with training performance. This behavior highlights the greater difficulty of achieving stable generalization when training data are scarce, as the model becomes more sensitive to the specific data partition and the patterns it learns. For the quantum models, a smoother decrease in the loss curves can be observed across most epochs for both training and validation sets. This behavior, together with the aforementioned difference in the number of epochs, indicates that the optimization process progresses more gradually than for the classical models. In the quantum model trained on the small dataset plot, both training and validation accuracy increases gradually, while the corresponding loss values decrease over a large number of epochs. Notably, in this particular run, the validation accuracy remains consistently higher and the validation loss lower than the corresponding training metrics for most of the training process. This behavior is atypical and is therefore treated as an anomalous observation specific to this run, rather than as a general characteristic of the model. The other final evaluation runs for the small dataset exhibit more typical training–validation behavior. In contrast, the quantum model trained on the large dataset shows a more pronounced initial adjustment in both accuracy and loss, followed by a more gradual convergence, with the training and validation curves remaining relatively close throughout most of the process.

\section{Discussion and conclusions}
\label{sec:discussion_and_conclusions}

\subsection{Analysis of advantages and capabilities}
\label{subsec:advantage_analysis}

The developed classical models are specifically designed to mimic the workflow of their quantum counterparts, yet they still achieve higher scores for the given metrics across two of the three employed datasets. However, it is important to consider the substantial differences in model complexity when comparing both approaches. The best-performing classical architecture for the small and large datasets consists of four stacked simple CRNN layers, comprising 15,930 and 118,439 trainable parameters, respectively. The large increase in the latter is due to the additional recurrent units in the first layers, which greatly increase the number of weight matrices and recurrent connections, and therefore the total number of trainable parameters. In contrast, the proposed quantum model requires at most 139 trainable parameters for the medium dataset, whereas the best-performing model for the small dataset contains only 94 parameters. Consequently, the quantum approach achieves competitive performance while using several orders of magnitude fewer trainable parameters than its classical counterpart.

A similar contrast can be observed in the memory requirements for both models. The classical architecture relies on recurrent hidden states with a dimensionality of up to 256 units per layer, whereas the quantum implementation operates with a maximum of only six qubits. Although current quantum hardware still suffers from important limitations related to noise, decoherence, and scalability, these results suggest that quantum recurrent models can represent sequential information with remarkably compact parameterizations. This parameter efficiency may become increasingly relevant as quantum hardware continues to mature, potentially enabling competitive models with significantly reduced memory and optimization requirements.

Training efficiency, measured by the number of epochs required for convergence, and generalization capability are also key aspects to consider. The results show that quantum models require substantially more epochs than their classical counterparts. Although both model types require the most epochs on the medium dataset, which indicates a clear capacity to learn more complex patterns at this size, the substantial gap in absolute epoch counts underscores a key limitation of the quantum implementation. The quantum models not only converge more slowly but also show far greater run-to-run variability, likely reflecting challenges related to weight initialization, seed randomness, and optimization stability. This indicates that, although the quantum approach achieves competitive predictive performance with far fewer trainable parameters, it does so at the cost of slower and less stable convergence. The reduced parameter count therefore does not translate into faster training in terms of epochs, and the classical models still exhibit more consistent generalization, especially on the larger and more complex dataset.

The convergence curves suggest that exploring alternative hyperparameter configurations, including more extreme values, could potentially improve the balance between model capacity and generalization performance, especially for the quantum implementation. However, the computational cost of such exploration is considerably higher, as each hyperparameter configuration requires 25 training executions due to five-fold cross-validation repeated five times. Combined with the longer training time required by the quantum model, this constraint limited the range of hyperparameter values considered during optimization. In any case, this does not affect the final results, since final performance is reported on an independent test set.

Although the quantum implementation achieves performance comparable to (in some cases exceeding) that of the classical model across the evaluated datasets, a key factor should be considered when interpreting these results: as stated in Section \ref{subsec:RNN_class}, the classical architecture was intentionally adapted to follow the same workflow as the quantum model, ensuring a fair comparison between both paradigms. Consequently, it does not necessarily represent the highest performance achievable with conventional ML techniques. 

It is also important to note that the quantum experiments were conducted in emulation, since NISQ hardware is still limited by noise, decoherence, and hardware constraints, making the execution of these models on real quantum devices challenging and potentially less reliable. Despite these limitations, the quantum approach presents a notable advantage in terms of model efficiency. With substantially fewer trainable parameters and less memory than the classical model, it achieves competitive classification performance. This indicates an efficient use of model capacity and suggests that, as quantum hardware continues to mature, these models may become an increasingly attractive alternative for ML applications.

\subsection{QML: Opportunities and limitations}
Recent advances in QML have positioned the field as one of the most active research areas of QC and artificial intelligence. During the last few years, significant progress has been achieved in both theoretical developments and practical implementations of quantum models. Hybrid quantum–classical architectures, variational quantum circuits, and quantum neural networks have become increasingly common approaches for tackling supervised learning, optimization, and generative modeling tasks \citep{tian2023qnn}. Although current applications are still limited in scale, the rapid development of quantum processors suggests that the capabilities of QML models may continue to improve considerably in the coming years.

These properties could enable more compact representations of complex data distributions, improved optimization procedures, and computational speedups for certain learning tasks. In particular, variational quantum circuits and quantum neural networks have been shown to achieve competitive or superior performance with substantially fewer trainable parameters than comparable classical models, thanks to the high expressivity afforded by superposition, entanglement, and interference \citep{gokhale2026review}. This parameter efficiency, demonstrated in Section \ref{subsec:advantage_analysis}, is especially relevant in high-dimensional settings. There, quantum kernels and quantum feature maps can implicitly operate in exponentially large Hilbert spaces, capturing non-linear correlations that are costly to represent classically \citep{yano2021featuremap}. Beyond QML, several quantum algorithms provide well-known theoretical speedups in related computational tasks. Shor’s algorithm, for instance, solves integer factorization and discrete logarithms in polynomial time, illustrating the potential for substantial reductions in time complexity when analogous subroutines are incorporated into learning pipelines \citep{ekert1996shor}. Although no general-purpose QML algorithm with comparable speedups has yet been established, it remains plausible that future advances will yield QML methods exhibiting similar advantages, motivating continued research into the integration of QC techniques within modern ML pipelines.

Despite this theoretical promise, QML currently faces several fundamental challenges that hinder its practical deployment on current hardware. A central issue is the presence of barren plateaus, where the gradients of the cost function vanish exponentially with the number of qubits, making training increasingly difficult as circuits grow in size \citep{mcclean2018barren}. This problem is compounded by the high cost of gradient estimation in quantum circuits, which typically relies on the parameter-shift rule and requires multiple circuit evaluations per parameter, leading to a substantial measurement overhead \citep{schuld2019gradient}. In addition, efficient state preparation remains a bottleneck, as encoding classical data into quantum states often demands deep circuits or specialized oracles that are not readily available on near-term devices \citep{agliardi2025encoding}. Together, these limitations constrain the expressivity and scalability of QML models and motivate the hybrid strategies discussed below.

As stated in Section \ref{sec:background}, QML faces important limitations due to the characteristics of existing quantum hardware since the field is currently situated within the NISQ era \citep{lau2022NISQ}. These constraints restrict the depth and complexity of quantum circuits that can be reliably executed, often limiting the practical performance of current QML models. As a consequence, many proposed approaches remain dependent on hybrid architectures where classical computation still performs a substantial portion of the workload. Nevertheless, ongoing advances in error mitigation, qubit stability, and quantum hardware scalability are expected to progressively reduce these limitations \citep{terhal2015error}. Future generations of fault-tolerant quantum computers could significantly expand the applicability of QML methods, potentially enabling models capable of addressing problems that are currently infeasible for classical approaches alone. Generally, estimating probabilities or expectation values from quantum measurements requires a large number of shots, introducing an additional hyperparameter that is critical for stable convergence. However, the architecture proposed in this work significantly reduces the required number of shots per timestep for two primary reasons. First, the task is a binary classification, meaning the decision boundary can be reliably resolved even with the limited numerical precision associated with low shot counts. Second, the sequential nature of the model involves repeated measurements across multiple timesteps, which effectively provides an inherent sampling mechanism.

\subsection{Conclusions and future work}

In this study, we explored the application of QRNNs for the binary classification of sleep events in multivariate time series data. We presented a hybrid classical-quantum QRNN architecture based on a variational quantum circuit and conducted a comprehensive comparative analysis against classical Machine Learning models based on RNNs.

Our results indicate that the quantum model achieves encouraging performance, with metrics that remain close to those of equivalent classical models and, in some cases, surpass them. This is achieved with a substantially smaller number of trainable parameters and lower computational requirements. Overall, these findings highlight the promising potential of Quantum Machine Learning, particularly QRNNs, for time-series analysis in medical applications and related fields.

Several research directions may be explored in the future to further improve the proposed approach and better assess the applicability of QML models. One possible extension involves the development of more complex QRNN architectures together with the evaluation of alternative quantum encoding strategies, which may allow richer representations of temporal information. In addition, further investigation of noise models and error mitigation techniques is required, particularly when deploying the models on real NISQ hardware, where decoherence and operational errors remain major limitations. Future research should systematically quantify the minimum shot budget required for the proposed quantum classifier, testing the hypothesis that shot counts can be substantially reduced without degrading binary classification accuracy.

Future studies could also evaluate the applicability of QRNNs to other types of physiological data and more challenging multi-class classification tasks, extending their use beyond the detection of apneic events. The integration of advanced classical preprocessing and feature engineering techniques with quantum models may also improve both stability and predictive performance. Finally, improving the interpretability of QRNN models represents a relevant challenge, since understanding how quantum-derived features contribute to classification decisions could provide valuable insights into both the learning process and the practical usefulness of quantum representations in biomedical applications.

\section*{Acknowledgements}

This study has been supported by projects \mbox{PID2023-147422OB-I00} and PID2024-159713OB-I00 funded by \mbox{MCIU/AEI/10.13039/501100011033} and by ERDF/EU, and project ED431F 2025/35 funded by Xunta de Galicia.

DAE received funding from project \mbox{RYC2022-038121-I}, funded by \mbox{MCIN/AEI/10.13039/501100011033} and European Social Fund Plus (ESF+). AMR and JDV have received support for the development of this article from ``Conseller\'ia de Educaci\'on, Ciencia, Universidades y Formaci\'on Profesional'' and ``Agencia Gallega de Innovaci\'on'' through ``Programa de ayudas a la etapa predoctoral'' \mbox{ED481A-2025/084} and \mbox{IN606A-2023/011} respectively. Both are researchers on a pre\-doctoral contract awarded by Xunta de Galicia.

The authors wish to acknowledge the support received from Centro de Investigación de Galicia ''CITIC'', a center accredited for excellence within the Galician University System and a member of the CIGUS Network, which receives subsidies from the Department of Education, Science, Universities, and Vocational Training of the Xunta de Galicia, and it is co-financed by the EU through the FEDER Galicia 2021--27 operational program \mbox{ED431G 2023/01}.

This research project was made possible through the access granted by the Galicia Supercomputing Center (CESGA) to its supercomputing infrastructure. The supercomputer FinisTerrae III and its permanent data storage system have been funded by the NextGeneration EU 2021 Recovery, Transformation and Resilience Plan, ICT2021-006904, and also from the Pluriregional Operational Programme of Spain 2014-2020 of the European Regional Development Fund (ERDF), ICTS-2019-02-CESGA-3, and from the State Programme for the Promotion of Scientific and Technical Research of Excellence of the State Plan for Scientific and Technical Research and Innovation 2013-2016 State subprogramme for scientific and technical infrastructures and equipment of ERDF, CESG15-DE-3114.

The Qmio Quantum Computing infrastructure has been funded from the European Union, through the Operational Programme Galicia 2014-2020 of ERDF\_REACT EU, as part of the European Union’s response to the COVID-19 pandemic.

The Sleep Heart Health Study (SHHS) was supported by National Heart, Lung, and Blood Institute cooperative agreements U01HL53916 (University of California, Davis), U01HL53931 (New York University), U01HL53934 (University of Minnesota), U01HL53937 and U01HL64360 (Johns Hopkins University), U01HL53938 (University of Arizona), U01HL53940 (University of Washington), U01HL53941 (Boston University), and U01HL63463 (Case Western Reserve University). The National Sleep Research Resource was supported by the National Heart, Lung, and Blood Institute (R24 HL114473, 75N92019R002).

The authors also wish to acknowledge Samuel Magaz Romero for his collaboration and contributions of ideas to the development of the proposed methodology.

\section*{Additional information}

 The authors declare no competing interests or conflicts of interest.

\section*{Data availability}

 The employed datasets are not publicly available due to the data access and use restrictions imposed by the Sleep Heart Health Study. However, the raw data (polysomnography files) are available upon request at the National Sleep Research Resource  through the following link: \url{https://sleepdata.org/datasets/shhs}.

\section*{Author contributions} 

Contributions stated in CRediT standard. AMR: Data Curation (Lead); Writing – Original Draft (Lead); Methodology (Equal); Software (Equal); Investigation (Equal). JDV: Writing – Original Draft (Supporting); Methodology (Equal); Software (Equal); Investigation (Equal). DAE: Resources (Lead); Investigation (Supporting); Supervision (Equal). EMR: Supervision (Equal); AG: Supervision (Equal); All authors: Writing – Review and Editing (Equal).

\bibliographystyle{unsrtnat}
\bibliography{references}  

\end{document}